\pdfoutput=1

\documentclass[11pt]{article}

\usepackage{ACL2023}

\usepackage{times}
\usepackage{latexsym}

\usepackage[T1]{fontenc}

\usepackage[utf8]{inputenc}

\usepackage{microtype}

\usepackage{inconsolata}

\usepackage{graphicx}

\usepackage{multirow}

\usepackage{eso-pic}

\title{Negated Complementary Commonsense using Large Language Models}

\author{Navid Rezaei \\
  ECE, University of Alberta \\
  Edmonton, Canada T6G 1H9 \\
  \texttt{nrezaeis@ualberta.ca} \\\And
  Marek Z. Reformat \\
   ECE, University of Alberta \\
  Edmonton, Canada T6G 1H9 \\
  University of Social Sciences\\ 
  90-113 \L{}\'{o}d\'{z}, Poland \\
  \texttt{reformat@ualberta.ca} \\}
  
  \AddToShipoutPicture*{%
  \AtPageUpperLeft{%
    \setlength\unitlength{1in}%
    \hspace*{\dimexpr\paperwidth-1in\relax}%
    \makebox[0pt][r]{%
      \raisebox{\dimexpr-\headheight-\headsep\relax}{%
        \parbox{\textwidth}{\centering\textbf{Appeared in Natural Language Reasoning and Structured Explanations Workshop (NLRSE)}\\ \textbf{ACL 2023}}%
      }%
    }%
  }%
}

\begin{document}
\maketitle
\begin{abstract}
Larger language models, such as GPT-3, have shown to be excellent in many tasks. However, we demonstrate that out-of-ordinary questions can throw the model off guard. This work focuses on finding answers to negated complementary questions in commonsense scenarios. We illustrate how such questions adversely affect the model responses. We propose a model-agnostic methodology to improve the performance in negated complementary scenarios. Our method outperforms few-shot generation from GPT-3 (by more than 11 points) and, more importantly,
highlights the significance of studying the response of large language models in negated complementary questions. The code, data, and experiments are available under: \url{https://github.com/navidre/negated_complementary_commonsense}.

\end{abstract}

\section{Introduction}
\label{sec: introduction}

\begin{figure}
    \centering
    \includegraphics[width=0.5\textwidth]{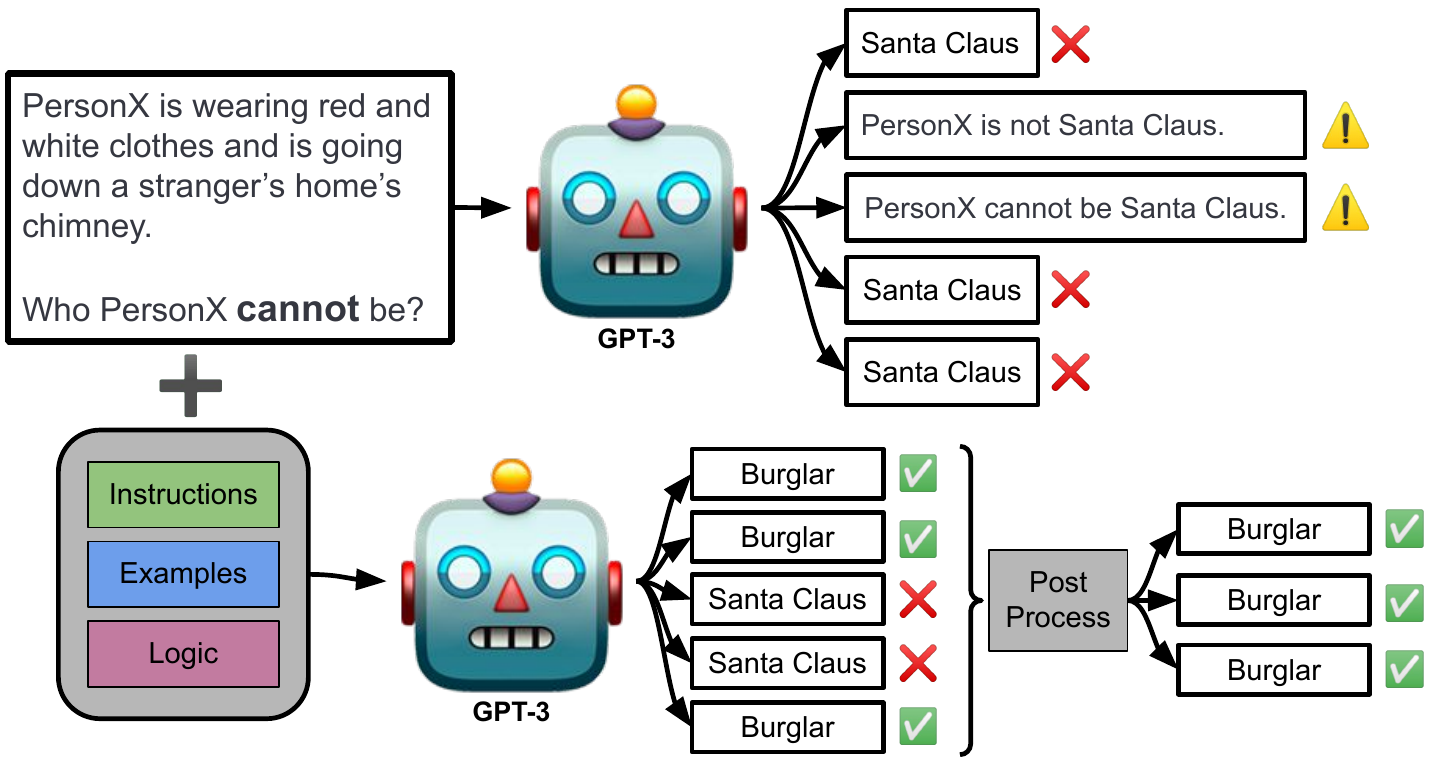}
    \caption{An example of a large language model (GPT-3) generating negated commonsense. Five responses per query are demonstrated. The applied pre-processing and post-processing can improve the performance of the models in negated commonsense cases. Non-specific answers, such as \textit{not Santa}, are considered incorrect.}
    \label{fig: main}
\end{figure}

The larger the language models (LLMs) become, the better they demonstrate new, outstanding capabilities. For example, one is conducting a conversation about commonsense scenarios. 
However, our interaction with LLMs has led us to observe that the models tend to emphasize the normal flow of events and seem to 
struggle with questions involving a negated form of verbs, such as \textit{not} or \textit{cannot}. An example of that is in Figure~\ref{fig: main}. 
Therefore, in this paper, we focus on demonstrating the issue and then suggest an approach to remedy the problem.

\begin{figure}
    \centering
    \includegraphics[width=0.28\textwidth]{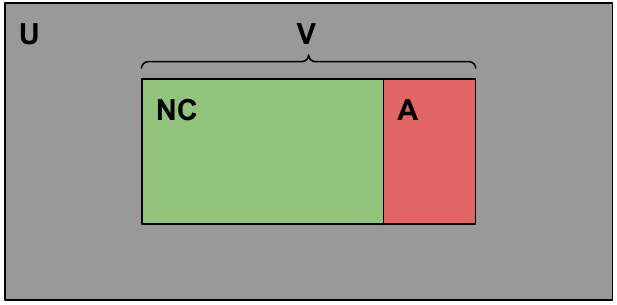}
    \caption{Venn diagram of answer sets: $U$ is the universal set of answers; $V$ is the set of all valid answers that includes two sets --  correct answers to a standard question $A$, and correct answers to its negated complementary version $\mathit{NC}$.}
    \label{fig: venn_diagram}
\end{figure}

To better clarify the problem statement, we start with an example and then formalize it using elements of the set theory. Let us look at the scenario in Figure~\ref{fig: main}; the standard question is ``Who PersonX can be?''. The answer to this question is \textit{Santa Claus}. The answer to the \textit{negated complementary} question -- ``Who PersonX \textit{cannot} be?'' -- should be all valid answers which are not the answer to the standard (can be) question. A valid answer fits the scenario described. In this case, we ask about a person, so a non-person cannot be a valid answer. To better illustrate the concept of a \textit{negated complementary} question, we refer to the basic notion of the complement of a set, Figure~\ref{fig: venn_diagram}. 
Furthermore, we define a set of correct answers to a \textit{negated complementary}, Equation~\ref{eq: nc}.
\begin{equation}
\mathit{NC} = V \cap A' = \{x \: | \: x\in V \land x\notin A\}
\label{eq: nc}
\end{equation}
where $\mathit{NC}$ represents answers to the \textit{negated complementary} question, $V$ is the set of all valid answers, $A$ is the set of correct answers to the standard question, and $A'$ is the complement of $A$ under the universal set of all answers ($U$). 

We focus our efforts on commonsensical questions as the uncertainty of results depends on the context and experiences of people answering the questions. As defined in \citep{lecun2022path}, commonsense is a collection of world models representing what is likely, plausible, or impossible. In light of that, our goal is to assess the ability of LLMs to answer plausible questions that could be refuted or accepted in a given context.

Given their pre-training nature, we hypothesize that LLMs have an inherent bias towards likely scenarios, which are the most repeated in the common text. Most of the text available on the web contains information supporting answers to `positive' questions, like, how to do things or where to go, not to questions such as how things could not be done or where not to go. It results in an imbalance of the training datasets due to the sparsity of plausible or impossible scenarios. In this paper, we demonstrate that LLMs have difficulty answering \textit{negated complementary} questions, which results in responses representing plausible, but not impossible, answers. Although LLMs are shown to have this shortcoming, we claim that enough instructions and examples, especially showing reasoning processes, can guide the LLMs into the right path to answer \textit{negated complementary} questions with commonsense context.

Our contributions are as follows. (1) We present an analysis exposing the shortcomings of LLMs when it comes to  \textit{negated complementary} questions in commonsensical scenarios. (2) We propose a novel methodology to improve the performance of the GPT-3 model when  \textit{negated complementary} questions are asked; compare the results with the results obtained using conventional methods. Our code, human-evaluation process, and data will be publicly available.


\section{Related Work}
\label{sec: related_work}
Language models with transformer architectures have revolutionized the natural language processing landscape in recent years \citep{Vaswani2017NIPS, devlin-etal-2019-bert}. It is shown that improved performance and new capabilities emerge when scaling up the size of language models \citep{brown2020language, chowdhery2022palm}, although more is needed in challenging tasks, such as commonsense \citep{rae2021corr}.

A body of research focuses on analyzing and extracting commonsense from language models \citep{west-etal-2022-symbolic, rezaei2022symmetry, Hwang2021COMETATOMIC2O, da2021analyzing}. Authors of \citep{jiang-etal-2021-im} focus on implications of negated statements and contradictions, where in a commonsense triple relationship (head-relation-tail), the head is either contradicted or logically negated. Comparably this paper focuses on negating relations instead of the head, as explained in Section~\ref{sec: methodology}. 

\section{Commonsense Data}
\label{sec: commonsense_data}

The commonsense dataset used in this paper is the ATOMIC-2020 dataset \citep{Hwang2021COMETATOMIC2O}. It includes general purpose commonsense knowledge, divided into three main categories -- physical, event-centered, and social commonsense. The ATOMIC 2020 dataset is licensed under CC-BY and we use it according to the license.


In our experiments, ten relation types are selected from the twenty-three relations from the ATOMIC-2020 dataset. These ten relation types performed worse in our initial evaluation of 
\textit{negated complementary} questions. The relations are: \textit{xWant}, \textit{xReact}, \textit{oWant}, \textit{CapableOf}, \textit{Desires}, \textit{HinderedBy}, \textit{isBefore}, \textit{isAfter}, \textit{AtLocation}, \textit{HasSubEvent}. The worse-performer triples are intuitively more common in the normal format in written language than their negated complementary versions, which can result in unbalanced training data.

The dataset is formatted in a triple style. Each atomic piece of data contains $\langle head - relation - tail \rangle$. For example, $\langle a \ curved \ yellow \ fruit \ (head) - CanBe \ (relation)
- banana \ (tail) \rangle$.

\section{Methodology}
\label{sec: methodology}

We propose a pipeline system to improve the performance on \textit{negated complementary commonsense} questions. The pipeline consists of an input prompting technique and a post-processing module. The input prompt adds relevant context and logic in the form of chain-of-thought prompting \citep{wei2022chain} to improve the LLM performance. The post-processing module selects the outputs with a higher chance of correctness and filters out the rest.

\subsection{Generating Negated Complementary Questions}
\label{sec: question_format}

As described in Section~\ref{sec: commonsense_data}, the used dataset is in the format of triples. To form a standard question, we use the head and the relation nodes and leave out the tail to be answered. By standard, we mean utilizing the head, relation, and tail, without any modifications. Assuming a triple, \textit{a curved yellow fruit} (head), \textit{CanBe} (relation), \textit{banana} (tail), the standard question is \textit{What can be a curved yellow fruit?}. The \textit{negated complementary} question is formed by negating the relation and verbalizing the resulting triple in question format: \textit{What cannot be a curved yellow fruit?} A valid answer to the standard question is \textit{banana}, and a reasonable response to the \textit{negated complementary} question is \textit{apple}. The process is visualized in Figure~\ref{fig: question_generation}. For the complete list of triple verbalizations, please see Appendix~\ref{app: verbalizations}.

\begin{figure}
    \centering
    \includegraphics[width=0.45\textwidth]{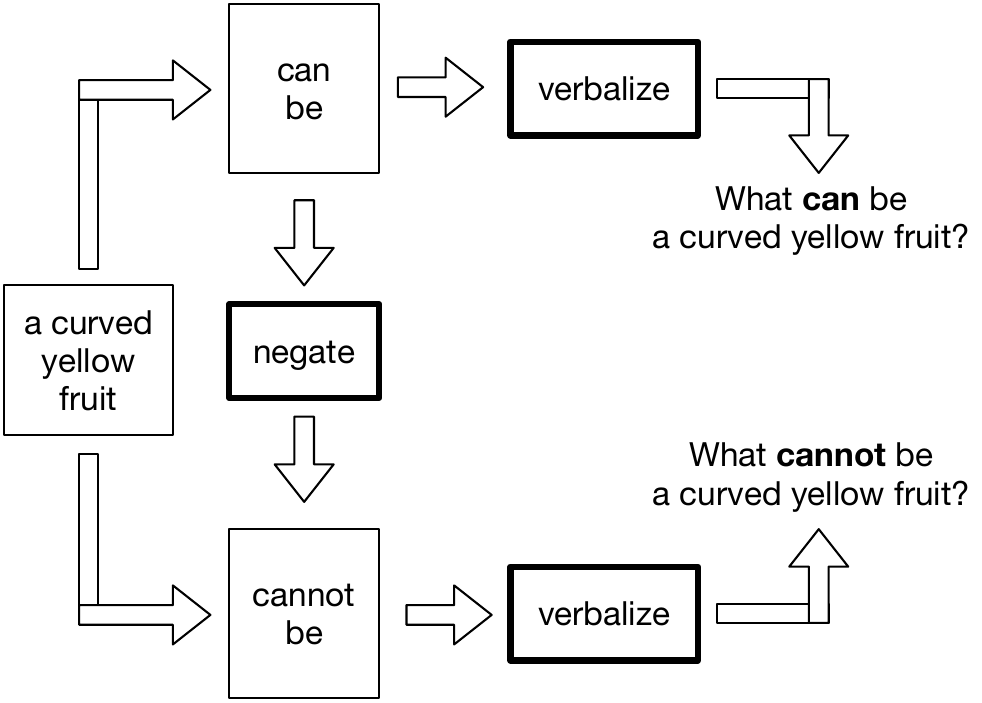}
    \caption{The process to automatically generate negated complementary questions from dataset triples. The head and relation nodes are used to form a question.}
    \label{fig: question_generation}
\end{figure}

\subsection{Prompting Technique}
\label{sec: prompt}

\begin{figure*}
    \centering
    \includegraphics[width=0.81\textwidth]{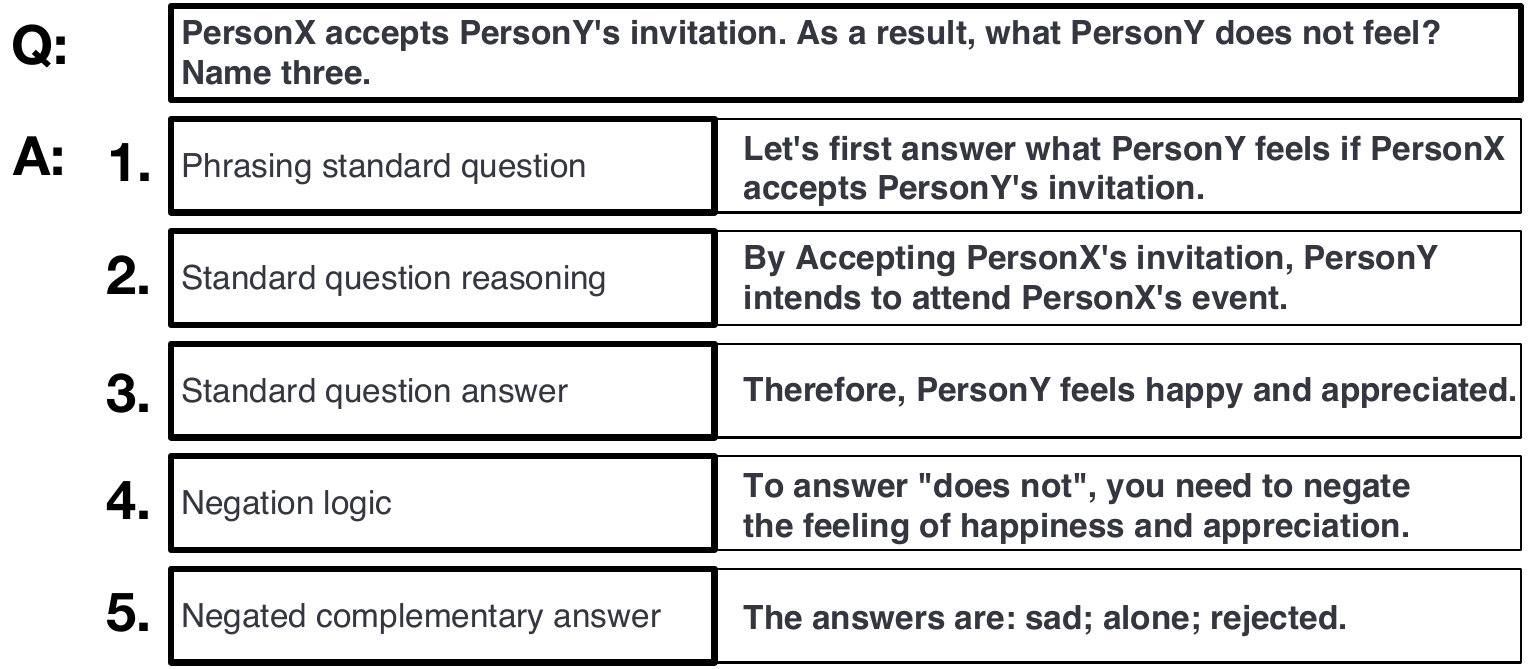}
    \caption{Chain-of-thought steps for each answer. The process is to answer the standard question first and then lead the model to answer the negated complementary version.}
    \label{fig: cot_steps}
\end{figure*}

The proposed methodology to improve the performance of LLMs relies on building an adequate prompt. It starts with a general introduction of what negations are and 
emphasizes a need to pay special 
 attention to the word \textit{Not}. The chain-of-thought prompt in each answer has five sections in sequence: 1) phrasing standard question; 2) standard question reasoning, 3) standard question answer; 4) negation logic, and 5)  \textit{negated complementary} question answer. The steps are visualized in Figure~\ref{fig: cot_steps}. For a fair comparison, we used the same number of five question/answer examples in the prompts. We also used the same questions for all prompts.

\subsection{Post Processing}
\label{sec: post_process}
Inspired by \citep{Kadavath2022LanguageM}, we feed the question and answer pair back to the GPT-3 model and ask if it considers a question/answer pair correct. The prompt has instructions for assessing an answer and includes five sample questions/answer pairs. Interestingly, this extra step can improve the results by almost one percent. To better understand the effect of this step, please refer to Table~\ref{tab: ablation}.

\section{Experiments}
\label{sec: experiments}

Experiments are conducted on each type of relation mentioned in Section~\ref{sec: commonsense_data}. A hundred data points (triples) are sampled randomly from the dataset. 
The head and relation from each triple are verbalized and fed into the GPT-3 model (\textit{text-davinci-002}). The goal is to predict the tail for two forms of questions: (1) standard question; (2) \textit{negated complementary} question. For each question, three responses are requested from the model. They are then parsed, and the answers (tails) are automatically extracted. Therefore, three possible tails are obtained for each head and relation, which results in $600$ total answers per method.

In social commonsense scenarios, PersonX and PersonY are used in place of gender-specific pronouns to make the questions and answers gender-neutral.

The experiments are done using the GPT-3 model \citep{brown2020language} with version \textit{text-davinci-002}, which has 175~billion parameters. The temperature is set to $0.7$, and in case of no answer, it is increased to $1.0$. The maximum length of the output is set between $100$ and $150$ tokens, depending on the method. The presence and frequency penalties are set to $0$. GPT-3 is commercially available, and we have used it within its intended usage and terms of service.

\subsection{Human Evaluations}
\label{sec: human_evaluations}
We use Amazon mTurk evaluations via AWS SageMaker to evaluate the results. Each answer is written in a sentence format and given to nine different annotators for assessment. Instructions and examples are provided with each question to assist the annotators better. The options to choose from are: (1) Makes sense; (2) Sometimes makes sense; (3) Does not make sense or incorrect; (4) The first part and the second part are not related; or not enough information to judge; (5) Unfamiliar to me to judge. The first two options are considered correct, the second two are considered incorrect, and the last is considered unfamiliar. To measure inter-rater reliability, we use Krippendorff's alpha and make sure the value is above acceptable amounts (minimum 0.667) \citep{Hayes2007AnsweringTC}. The evaluators were paid based on AWS guidelines.

\begin{table*}[h]
\centering
\begin{tabular}{lll}
\hline
\textbf{Method} & \textbf{Standard} & \textbf{Negated Complementary}\\
\hline
Few-shot & \textbf{88.7\%} & 78.7\% \\
Ours &  88.1\% & \textbf{89.8\%} \\
\hline
\end{tabular}
\caption{Our method compared with the few-shot method when applied to ATOMIC-2020 dataset.}
\label{tab: comparison}
\end{table*}

\subsection{Results}
\label{sec: results}
As seen in Table~\ref{tab: comparison}, our method outperforms the few-shot method by more than eleven percentage points when answering \textit{negated complementary} questions. The few-shot method includes five different questions in the prompt with their answers without chain-of-thought prompting. The performance of our method can mainly be attributed to the specific chain-of-thought prompting with negation logic description, Figure~\ref{fig: cot_steps}. More information about the main contributing factors is in Section~\ref{sec: ablation}. Although chain-of-thought prompting seems to help the \textit{negated complementary} questions, it adversely affects answers to the standard questions. Please note that the chain-of-thought prompt for the standard questions does not include negation logic, and a post-processing technique similar to negated complementary questions is performed.

\subsection{Ablation Studies}
\label{sec: ablation}

To gain insight into the importance of elements of our method, we perform an ablation study, Table~\ref{tab: ablation}. As we can see, adding standard question reasoning (step 2 of Figure~\ref{fig: cot_steps}) results in more than 7\% improvement in the results. Adding the thought process explaining the negation logic (steps 1, 3, and 4 of Figure~\ref{fig: cot_steps}) adds another 3\% performance improvement. Finally, the post-processing (Section~\ref{sec: post_process}) is responsible for about 1\% improvement in the results.

\begin{table}[h]
\centering
\begin{tabular}{ll}
\hline
\textbf{Method} & \textbf{Neg. Comp.}\\
\hline
Ours & 89.8\% \\
Ours-wo-pp & 89.0\% \\
Ours-wo-nl-pp & 86.0\% \\
Few-shot & 78.7\% \\
\hline
\end{tabular}
\caption{Ablation study of the method: \textit{Ours-wo-pp} is ours without post-processing; \textit{Ours-wo-nl-pp} is ours without negation logic and post-processing.}
\label{tab: ablation}
\end{table}


\section{Conclusions}
In this paper, we demonstrate how simple changes in question formats, which can be trivial for humans, can be challenging for large language models (LLMs). We specifically focus on \textit{negated complementary} questions in a commonsense context, which is constructed by negating a relation in a commonsense triple. Given the vast amount of knowledge embedded in LLMs, we show that by appropriate guidance, the models could perform well on \textit{negated complementary} tasks. Our method results in more than eleven percent improvement compared to the vanilla few-shot method. Given the widespread usage of LLMs and their growth rate, we believe focusing on and solving the model's weaknesses is imperative. As future work, \textit{negated complementary} task can be further analyzed in different formats, such as sentence instead of a question, and also different contexts, e.g., new datasets.

\section*{Limitations}
The experiments in this paper have focused on the \textit{negated complementary} task in the context of commonsense and the format of questions. However, it is interesting to experiment with other contexts, such as mathematical datasets and other formats, such as sentences instead of questions.

This paper only uses the English language in the \textit{negated complementary} task experiments, so further investigation is needed in other languages to understand better the limitations of large language models across other languages.

GPT-3 is commercially available, and the cost can be a limitation. For example, the current price for \textit{text-davinci-002} model is \$0.02 per 1,000 tokens.

\section*{Ethics Statement}
Given the widespread use of large language models and their growth, more software systems will depend on them. This could improve productivity and accessibility, but any vulnerability in large language models can propagate through the system and affect the end users. This work focused on distorted commonsense scenarios that are almost trivial for humans but can be challenging for large language models. Not only we highlighted the \textit{negated complementary} questions issue, but we also suggested practical solutions that do not require extensive computation. We believe this line of research can ultimately benefit end users in terms of productivity, reliability, and accessibility.

\bibliography{anthology,custom}
\bibliographystyle{acl_natbib}

\appendix

\section{Verbalizations}
\label{app: verbalizations}
The questions are verbalized from triples using pre-defined formats. Table~\ref{tab: verbalizations} summarizes the verbalizations organized by relation types. The question template formats are inspired by the sentence format used in \citep{Hwang2021COMETATOMIC2O}.

\begin{table*}
\centering
\begin{tabular}{ll}
\hline
\textbf{Relation} & \textbf{Question} \\
\hline
xWant & [head]. What does PersonX want to do? \\ 
\cline{2-2} & [head]. What does PersonX not want to do? \\
\hline
xReact & [head]. What does PersonX feel about it? \\ 
\cline{2-2} & [head]. What does PersonX not feel about it? \\
\hline
oWant & [head]. What does PersonY want to do? \\ 
\cline{2-2} & [head]. What does PersonY not want to do? \\
\hline
CapableOf & What is [head] capable of? \\ 
\cline{2-2} & What is [head] not capable of? \\
\hline
Desires & What does [head] desire to do? \\ 
\cline{2-2} & What does [head] not desire to do? \\
\hline
HinderedBy & [head]. What can hinder/obstruct it? \\ 
\cline{2-2} & [head]. What cannot hinder/obstruct it? \\
\hline
isBefore & [head]. What happens before it?\\ 
\cline{2-2} & [head]. What does not happen before it? \\
\hline
isAfter & [head]. What happens after it? \\ 
\cline{2-2} & [head]. What does not happen after it? \\
\hline
AtLocation & Where is the [head] located? \\ 
\cline{2-2} & Where is the [head] not located? \\
\hline
HasSubEvent & [head]. What will you do while: [head]? \\ 
\cline{2-2} & [head]. What you will not do while: [head] \\
\hline
\end{tabular}
\caption{Question templates for each relation type. The first row for each relation shows the standard question format, and the second row shows the negated complementary format. [head] refers to the head in a triple.}
\label{tab: verbalizations}
\end{table*}

\section{Human Evaluation Instructions}
\label{sec: instructions}
The following instructions are given to each human evaluator to better understand and respond to the task:

Based on your own commonsense, choose one of the five options. Examples are provided in the description. IMPORTANT: Please note the CANNOT, DO Not, and other negated cases.

Instruction notes:
Based on your own commonsense, choose one of the five options. Examples are provided in the description. 

IMPORTANT: Please note the CANNOT, DO Not, and other negated cases.

\begin{enumerate}
  \item Instead of names, PersonX and PersonY are used to be gender-neutral.
  \item Please ignore grammatical errors and focus on commonsense.
  \item If a response is vague, such as \textit{not fireman}, or if a random word does not fit the scenario, please choose 4 (not enough information).
\end{enumerate}

Added to the instructions, we also provided some examples to clarify the task better:

\textbf{Unfamiliar to me to judge:} PersonX discovers a new planet. The planet is in the Alpha Centauri system.

\textbf{First part and second part are not related! Or not enough information to judge:} PersonX rides a bike. Elephants are not birds. (Although the second part is correct, it is not related to the first part)

\textbf{Makes sense:} It is NOT likely to see an elephant on the table.

\textbf{Does not make sense:} It is likely to see an elephant on the table.

The answers are written in a text format to make it easier for human evaluators. Here is an example:. Here is an example:

PersonX sends PersonY to the showers. PersonX will not be amused.

\section{ChatGPT}
\label{sec: chat_gpt}
The experiments of this paper were done using \textit{text-davinci-002} version of GPT-3. The new version of GPT-3 with reinforcement learning from human feedback (RLHF) is introduced and is called ChatGPT. Out of curiosity, we did limited experiments with ChatGPT to see if the problem of \textit{negated complementary} questions persists. Here are a few answers from ChatGPT to the question shown in Figure~\ref{fig: main}:

Q: PersonX is wearing red and white clothes and is going down a stranger’s home’s chimney. Who PersonX cannot be?

A: It is likely that PersonX is not Santa Claus, as he is commonly depicted wearing red and white clothes and going down chimneys.

As we can see, the model answers with \textit{not Santa Claus} and then describes Santa Clause. Asking the same question again, we get the following answer:

A: PersonX cannot be a real person as it is impossible for someone to physically fit into a chimney. The scenario you described is likely referencing a fictional character, such as Santa Claus, who is often depicted as wearing red and white clothes and entering homes through chimneys on Christmas Eve to deliver presents to children.

In this answer, there is a good start of context; however, somewhere within the text, the model forgets about the \textit{cannot} word and again describes Santa Claus. For the third time, we ask the same question and we get the following answer:

A: It is likely that PersonX is Santa Claus, a fictional figure who is often depicted wearing red and white clothes and delivering presents to children through the chimneys of houses.

The sentence is technically correct, but it is not what we asked for, which is who \textit{cannot} be. Therefore, we can consider the last answer incorrect as well.

Given the limited interaction with ChatGPT, it seems that the \textit{negated complementary} task issue persists.

\end{document}